\documentclass[conference]{IEEEtran}
\usepackage{paper_header}

\DeclareMathOperator{\Bl}{Bl}
\DeclareMathOperator{\Pl}{Pl}

\begin{document}

\title{Evidence Updating for Stream-Processing in Big-Data: Robust Conditioning in Soft and Hard Data Fusion Environments}

\author{%
\IEEEauthorblockN{Thanuka Wickramarathne~\IEEEmembership{Member,~IEEE,}}
\IEEEauthorblockA{Department of Electrical and Computer Engineering\\
University of Massachusetts Lowell, Lowell, MA 01852 USA\\
\ul{Email: thanuka@uml.edu}}}


\maketitle

\begin{abstract}
Robust \emph{belief revision} methods are crucial in \emph{streaming data} situations for updating existing knowledge (or beliefs) with new incoming evidence. \emph{Bayes conditioning} is the primary mechanism in use for belief revision in \emph{data fusion} systems that use probabilistic inference. However, traditional conditioning methods face several challenges due to inherent data/source imperfections in big-data environments that harness \emph{soft} (i.e., human or human-based) sources in addition to \emph{hard} (i.e., physics-based) sensors. The objective of this paper is to investigate the most natural extension of Bayes conditioning that is suitable for evidence updating in the presence of such uncertainties. By viewing the evidence updating process as a thought experiment, an elegant strategy is derived for robust evidence updating in the presence of extreme uncertainties that are characteristic of big-data environments. In particular, utilizing the \emph{Fagin-Halpern} conditional notions, a natural extension to Bayes conditioning is derived for evidence that takes the form of a general belief function. The presented work differs fundamentally from the \emph{Conditional Update Equation (CUE)} and authors own extensions of it. An overview of this development is provided via illustrative examples. Furthermore, insights into parameter selection under various fusion contexts are also provided.
\end{abstract}

\section{Introduction}

\bi{Overview.}
It's a streaming world---from financial markets to transportation to health monitoring to e-commerce applications, most of today's data are generated and received in real-time as streams~\cite{Val09}. With real-time processing bearing the promise of improved efficiency and creating new opportunities many application domains, \emph{stream processing}~\cite{Esp15,jams16,Li_ToBD16} has become the latest trend in the big-data world. The ability to predict future system states from real-time data streams while automatically accounting for data distribution drifts via a `single-pass' processing of data~\cite{Wang_PRL17} is a critical step in developing robust streaming processing methods for reasoning upon rapidly changing information. In particular, it is often required to `refine' existing knowledge (e.g., about a state of the system), commonly referred to as \emph{evidence updating} (or \emph{belief revision}) as new evidence is generated~\cite{Kir03}. To preserve the integrity of \emph{data fusion}~\cite{Hal92_book}, adequately accounting for numerous uncertainties~\cite{Wic12_ICBF} is paramount, especially in big-data environments where \emph{soft} (i.e., human or human-based) sources are frequently utilized in addition to \emph{hard} (i.e., physics-based) sensors. Among many uncertainty modeling/handling methodologies, notions of probability are very much likely to play a major role in big-data approaches as  %
(a) data are intrinsically probabilistic in nature and %
(b) ability of probabilistic approaches in reducing the size of the input data that is needed to be processed (by each machine) through randomization techniques. %
In this paper, we present a new evidence updating scheme that is derived as a natural extension of \emph{Bayes conditioning}, the primary belief revision mechanism utilized in probability theory~\cite{Lam94,Hec95,Fri97b, Jen01_book,Hua02,Man05}, to tackle the challenges associated with belief revision in such big-data environments.

\IEEEpubidadjcol 

\bi{Background.}
\emph{Conditioning}~\cite{Fag91_UAI,Ken92, Chr95} is the primary method for \emph{belief revision} in a vast majority of data fusion~\cite{Hal92_book,Abi92_book, Mitchell12_book} systems which employ probabilistic inferencing~\cite{Bog87,Pea88_book,Elo01, Del04,Che05, Ris05,Den06_SMC, Mar06, Mas11}. As new evidence becomes available, existing belief of the \emph{propositions} of interest (e.g., state of a sensor or agent,  knowledge-base (KB), etc.) are updated (or revised) to reflect the new evidence. In probability theory, the Bayes conditional accomplishes exactly this task. Let us consider the following example to illustrate the classical belief revision process.

\begin{example}[MVP Poll]
\label{ex:e_poll1}
Consider a sports news agency that maintains a KB of current voter preferences for the top five candidates $\Theta\equiv\{c_1, c_2, c_3, c_4, c_5\}$ for the most valuable player (MVP) award of a basketball league. The agency maintains an up-to-date KB via a probability mass function (pmf) $P_k(\cd)$, where $k$ is the discrete time index for current time $t_k$. The agency generates $P_k(\cd)$ by aggregating predictions generated by various sports analysts. Suppose the players $c_1$ and $c_2$ get injured early in the season. Now, the agency concludes with 100\% certainty that the two players will not play for the remainder of the season. Therefore, the KB is updated to reflect these changes by conditioning $P_k(\cd)$ with respect to (w.r.t.) the conditioning event $A=\Theta\setminus\{c_1,c_2\}=\{c_3,c_4,c_5\}$ thus yielding
\begin{equation}
\label{eqn:ex_1_bayes}
  P_{k+1}(c_i)
	= P_k(c_i\mid A)=\frac{P_k(c_i\cap A)}{P_k(A)},\; i=1,\ldots,5,
\end{equation} 
where $P_{k+1}(\cd)$ denotes the updated voter preference.
\end{example}

Evidence updating in fusion engines that utilize probabilistic inferencing is often carried out in a similar fashion to Example~\ref{ex:e_poll1}, where existing beliefs are conditioned w.r.t. an event $A$ that characterizes the changes to existing ``conditions.'' However, it is often difficult to characterize such changes via a \emph{single} event in complex sensing and fusion situations that are characteristic of big-data environments~\cite{Bur06}, especially when non-traditional sources of evidence (e.g., \emph{soft data} in the forms of witness reports, expert opinions, blogs, etc.) are also being used for gathering evidence~\cite{Hall12_book}. The evidence provided by such sources are often complex data structures that are difficult to represent via a single event with certainty or even via a probabilistic model. For instance, given the injuries to players $c_1$ and $c_2$ in Example~\ref{ex:e_poll1}, a sports analyst $\mc{A}^*$ may very well conclude that  \emph{``with a 75\% confidence, $c_1$ and $c_2$ will not return and in that case $c_3$ will most likely be the MVP"} (see Example~\ref{ex:e_poll2}).

\begin{figure}[htb!]
\vspace*{-0.1in}
\centering
\subfigure[hard evidence]{
  		\includegraphics[width=1.6in]{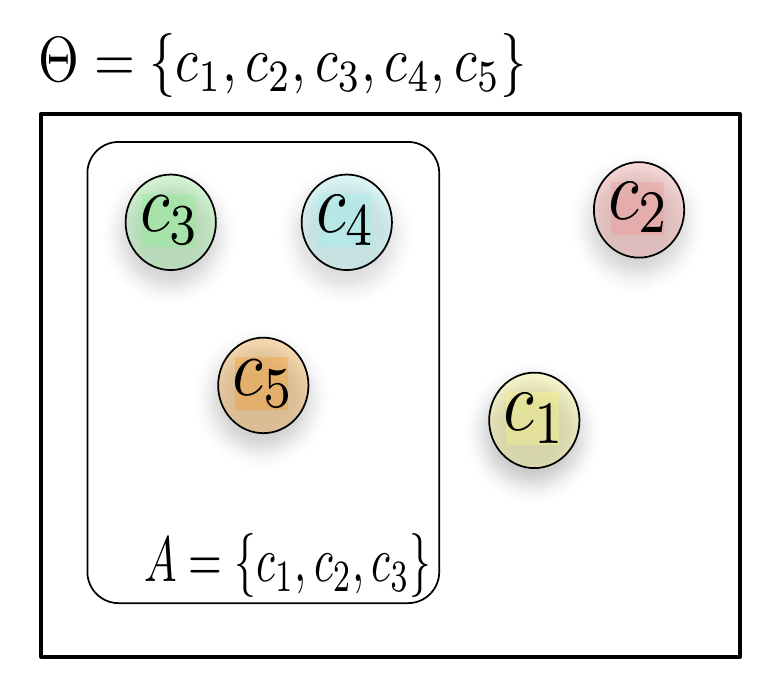}}\hfill
\subfigure[soft evidence]{
  		\includegraphics[width=1.6in]{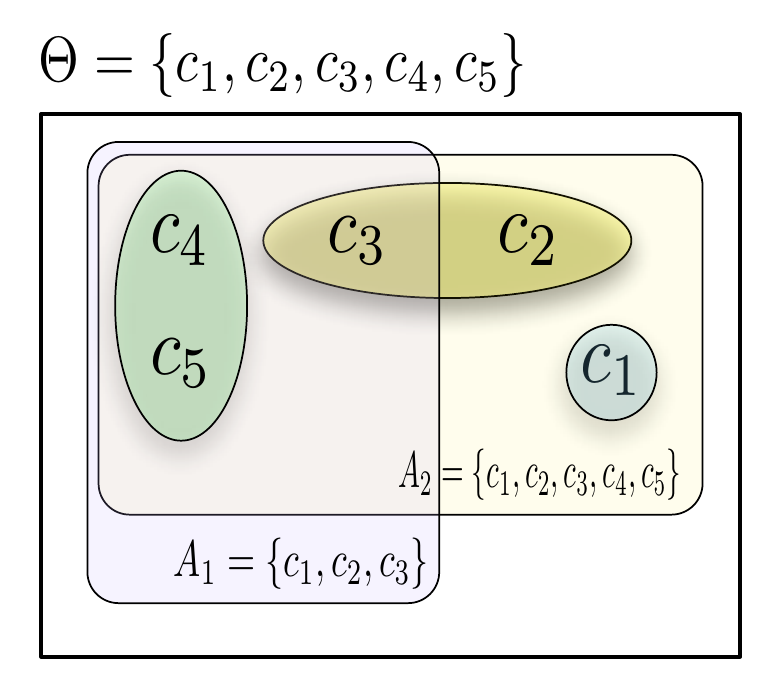}}
\vspace*{-0.05in}
\caption{Conditioning in big-data environments for MVP poll in Example~\ref{ex:e_poll1} and Example~\ref{ex:e_poll2}. Fig.~(a) shows the classical Bayes conditioning, where the conditioning event $A$ has occurred with certainty. Fig.~(b) shows a soft environment where preferences are assigned to `composite' sets (see Example~2) and, changes to existing conditions are captured via conditioning events $A_1$ and $A_2$ with 75\% and 25\% confidence, respectively.}
\label{fig:ex1}
\vspace*{-0.1in}
\end{figure}

\bi{Challenges.}
The difficulties associated with evidence updating in such complex environments is rooted in the types of data/source imperfections that we may encounter. Given the very nature of big-data (viz., subjective, qualitative, and unstructured), one may be reluctant to revise existing beliefs with complete certainty (w.r.t. a single conditioning event). Furthermore, one must also take into account the differences in data sources, such as the \emph{reliability} of the evidence source, \emph{credibility} of the evidence, and \emph{frame of discernment (FoD)} (i.e., the scope of source expertise). Purely probabilistic methods may pose several challenges in terms of adequately representing complex data uncertainties (that are characteristic of such data) in order to generate inferences that are robust against such data/source imperfections. Given the fact that vast majority of existing \emph{hard data} (i.e., data generated by conventional physics-based sensors) fusion and tracking systems are based on probabilistic methods, a complete transition to a new framework, which does not support probabilistic inferencing, may also not be a feasible option.

\bi{Contributions.}
By viewing the evidence updating process as a \emph{thought experiment,} we devise an elegant strategy for robust evidence updating in the presence of extreme uncertainties. With a belief theoretic~\cite{Sha76_book} core, our proposed method generalizes the belief theoretic \emph{Fagin-Halpern} conditional notions~\cite{Fag91_UAI}, thus allowing one to account for various data/source imperfections that are characteristic to complex soft/hard data fusion environments. Furthermore, a novel data fusion rule is derived as a natural extension of these ideas. The presented extension differs fundamentally from the previously published work on \emph{Conditional Update Equation (CUE)} as appeared in~\cite{Pre07, Pre09_ICIF}, including the authors own work in~\cite{Wic10_WTBF,Wic12_ICBF,Wic13_ICASSP}. However, it can be thought of as an extension of~\cite{Kul04}. In this paper, we provide an overview for the development of generalized conditional approach via illustrative examples. Then, we derive several algebraic and fusion properties of the new approach and compare them to the properties of CUE. Moreover, we provide insights into where each fusion rule may apply and also provide insights for parameter selection under various fusion contexts.

\section{Preliminaries}
\label{sec:prelims}

\paragraph{Basic Notions}
Let $\Theta\equiv\{\theta_1,\ldots,\theta_n\}$ denote the total set of mutually exclusive and exhaustive propositions. In DS theory, $\Theta$ is referred to as Frame of Discernment (FoD), where a proposition $\theta_i$ referred to as a \emph{singleton} represents the lowest level of discernible information. We use $|\Theta|$ and $2^{\Theta}$ to denote the cardinality and the power set of $\Theta$, respectively. Elements in $2^{\Theta}$ form all the propositions of interest in DS theory. When the FoD is clear from the context, we use $\ol{B}$ to denote all singletons in $\Theta$ that are not included in $B$; otherwise, we indicate these singletons via $\Theta{\setminus}B$. The `support' for proposition $B$ is provided via a \emph{basic probability assignment (BPA)} or \emph{mass assignment:}

\begin{definition}[BPA or Mass Assignment]
\label{def:bba}
The mapping $m: 2^{\Theta}\mapsto [0,1]$ is a BPA for the FoD $\Theta$ if $m(\emptyset)=0$ and $\sum_{B\subseteq\Theta} m(B)=1$.
\endproof
\end{definition}

A proposition receiving a positive BPA is referred to as a \emph{focal element;} the set of focal elements is the \emph{core $\mathcal{F}$;} the triple $\{\Theta, \mathcal{F}, m\}$ is the corresponding \emph{body of evidence (BoE).} The mass assigned to a proposition is free to move into the individual singleton objects that constitute the composite proposition thus generating the notion of \emph{ignorance.} Indeed, complete lack of evidence to discern among the propositions can be conveniently captured via the \emph{vacuous BoE} where the FoD $\Theta$ itself is the only focal element (i.e., $m(\Theta)=1$).

\paragraph{Belief and Plausibility}
While $m(B)$ measures the support assigned to proposition $B$ \emph{only,} the belief assigned to $B$ takes into account the supports for all proper subsets of $B$ as well. In other words, $\Bl(B)$ represents the total support that can move into $B$ without any ambiguity; and $\Pl(B)$ represents the extent to which one finds $B$ plausible. 

\begin{definition}[Belief and Plausibility]
\label{def:bdp}
For $B\subseteq\Theta$ in the BoE $\{\Theta, \mathcal{F}, m\}$, $\Bl: 2^{\Theta}\mapsto [0,1]$ where $\Bl(B)=\sum_{C\subseteq B} m(C)$ is the \emph{belief} of $B$; and $\Pl: 2^{\Theta}\mapsto [0,1]$ where $\Pl(B)=1-\Bl(\ol{B})$ is the \emph{plausibility} of $B$.
\endproof
\end{definition}

When each focal set contains only one element, i.e., $m(B)=0,\;\forall |B|\neq 1$, belief functions become probability functions. In such a case, the BPA, belief and plausibility all reduce to probability. A probability distribution $\tx{Pr}(\cdot)$ such that $\Bl(B)\leq\tx{Pr}(B)\leq\Pl(B),\,\forall B\subseteq\Theta$, is said to be \emph{compatible} with the underlying BPA $m(\cdot)$. An example of such a probability distribution is the \emph{pignistic probability distribution $\tx{BetP}(\cdot)$} \cite{Sme99_UAI}
\begin{equation}
  \tx{BetP}(\theta_i)
    =\sum_{\theta_i\in B\subseteq\Theta}
     \frac{m(B)}{|B|}.
\label{eqn:BP}
\end{equation}

\paragraph{Conditional Notions}
Let $\hat{\mf{F}}$ denote the set of propositions with non-zero belief, i.e., $\hat{\mf{F}}=\{B\subseteq\Theta\mid \Bl(B)>0\}$. The Fagin-Halpern (FH) conditional notions in DS theory \cite{Fag91_UAI} are applicable whenever the conditioning proposition $A$ belongs to $\hat{\mf{F}}$.

\begin{theorem}[Fagin-Halpern (FH) Conditionals] 
\label{thm:CBP}
\cite{Fag91_UAI}
For the conditioning event $A\in\hat{\mf{F}}$ and $B\subset\Theta$ in the BoE $\mc{E}=\{\Theta, \mf{F}, m\}$, the \emph{conditional belief $\Bl(B|A): 2^\Theta\mapsto [0,1]$} and the \emph{conditional plausibility $\Pl(B|A): 2^\Theta\mapsto [0,1]$ of $B$ given $A$} are
\begin{align*}
  \Bl(B|A)
    &=\frac{\Bl(A\cap B)}{\Bl(A\cap B)+\Pl(A\cap\ol{B})}; \\
  \Pl(B|A)
    &=\frac{\Pl(A\cap B)}{\Pl(A\cap B)+\Bl(A\cap\ol{B})},
\end{align*}
respectively.
\endproof
\end{theorem}

A proposition with positive mass after conditioning is referred to as a \emph{conditional focal element}. The collection of conditional focal elements that are generated with respect to the conditioning event $A$ is referred to as the \emph{conditional core} and denoted by $\mf{F}_{\Theta|A}$. Thus $\mf{F}_{\Theta|A}=\{B\subseteq\Theta\mid m(B|A)>0\}$, where $A\in\hat{\mf{F}}$ and $m(\cdot|A): 2^{\Theta}\mapsto[0,1]$ is the corresponding conditional BPA related to $\Bl(\cdot|A)$ via the M\"obius transformation \cite{Sha76_book}
\begin{equation}
  \label{eqn:bba_bel}
  m(B|A)
    =\sum_{C\subseteq B}(-1)^{|B-C|}\Bl(C|A),\;
     \forall B\subseteq\Theta.
\end{equation}

\paragraph{Evidence Updating}
This refers to the process of updating the evidence in a BoE $\mc{E}_k$ with evidence received from another BoE $\mc{E}_k^*$, to arrive at $\mc{E}_{k+1}$. Here $k$ denote the discrete update index. We denote this as $\mc{E}_{k+1}\equiv\mc{E}_k\lhd\mc{E}_k^*$.

\begin{definition}[Conditional Update Equation (CUE)]
\label{def:CUE}
\cite{Pre09_ICIF}
The CUE that updates $\mc{E}_k$ with the evidence in $\mc{E}_k^*$ is
\[
  Bl_{k+1}(B)
    =\alpha_k\,Bl_k(B)
       +\!\!\!\!
        \sum_{A\in \mf{F}_k^*} 
        \beta_k(A)\,Bl_k^*(B|A),
\vspace*{-0.1in}
\]
where the parameters $\alpha_k, \beta_k(\cd)\in\Re^+$ satisfy
\begin{math}
  \alpha_k+\sum\limits_{A\in\mf{F}_k^*}\beta_k(A)
    =1.
\end{math}
\endproof
\end{definition}

The CUE proposed in \cite{Pre09_ICIF} provides several interesting properties applicable to the task at hand. However, the updating mechanism presented in this paper fundamentally differers from CUE and previous work of the author. In fact, we provide a comprehensive discussion on these differences and how it affects the applicability and choice of parameters.

\section{Conditioning in Soft/Hard Fusion Environments}
\label{sec:cond_softhard}

As now being frequently harnessed in many big-data environments, soft data (i.e., human or human-generated) data plays a crucial role in inferencing tasks primarily due to their ability to provide complementary (to hard sources) and both critical and time-sensitive information. However, given the imperfect nature (subjective, incomplete, unstructured, inconsistent, contradictory, etc.) of these data/sources, one may not wish to sacrifice the integrity of the inferencing task by simply conditioning an existing KB w.r.t. soft evidence. Here, via a thought experiment, we develop a generalized conditioning operation as a direct extension of Bayes conditioning operation to account for these challenges.

\subsection{The Case of One Conditioning Event}
Let us look at a case where the new evidence comes in the form of an occurrence of one event, but not necessarily with 100\% certainty as in the traditional Bayesian conditioning.

\begin{example}[MVP Poll v.2]
\label{ex:e_poll2}
Suppose the sports agency in Example~\ref{ex:e_poll1} now maintains voter preferences via DS BoE $\mc{E}_k\equiv\{\Theta,\mf{F}_k,m_k(\cd)\}$ and is interested in updating its KB using the incoming evidence $\mc{E}^*_k$ 
from a regional ``star'' sports analyst $\mc{A}^*$. When $c_1$ and $c_2$ gets injured, $\mc{A}^*$ has now inferred that the two players will not return with 75\% confidence. Now, what is the best strategy to update $\mc{E}_k$ w.r.t. to recent changes?
\end{example}

Conditioning scenario in Example~\ref{ex:e_poll2} is clearly different and rather complicated from that of Example~\ref{ex:e_poll1}, where the existing KB was simply conditioned w.r.t. and event that characterizes the changes. Furthermore, the problem at hand is also not an evidence combination, since the agency is only interested in updating its voter preference KB using the newly acquired evidence from $\mc{A}^*$. This is a typical scenario in a soft/hard fusion network, where a node maybe interested in simply updating its existing state (or beliefs) by ``eavesdropping'' to evidence that is being relayed through it. 

\paragraph{The Probabilistic Case}
For illustration purposed, let us assume that the current voter preferences are stored in KB as $P_k(\cd)$ as in Example~\ref{ex:e_poll1}. 
In this case, when a conditioning event, such as $A=\Theta\setminus\{c_1,c_2\}=\{c_3,c_4,c_5\}$ (i.e., $c_1$ and $c_2$ are no longer in the running) is specified with 100\% certainty, one can interpret this as the original FoD being deflated (i.e., the \# of available candidates are being reduced) to $(c_3,c_4,c_5)$. Therefore, the conditioning operation, redistributes originally cast voter preferences, as given by 
\begin{equation}
\label{eq:p2}
  P_{k+1}(c_i)
	= P_k(c_i\mid A)=\frac{P_k(c_i\cap A)}{P_k(A)},\; i=1,\ldots,5.
\end{equation} 
Here, note that conditioning simply normalizes (such that they sum to 1) the originally cast voter preferences to propositions $c_3,c_4$ and $c_5$, as $P_k(c_1\cap A)=P_k(c_2\cap A)=0$.

Now, if the event $A=\Theta\setminus\{c_1,c_2\}=\{c_3,c_4,c_5\}$ only has 75\% confidence (or certainty) associated with it, how can one update the KB? While a philosophical discussion on generating the most precise interpretation of ``... $c_1$ and $c_2$ will not return with 75\% confidence" is out of the scope of this paper, it is clear from the context that $c_1$ and $c_2$ will not return with 75\% confidence does not necessarily mean that they they will return with 25\%. Then, as proposed in~\cite{Kul04} for updating in belief functions, one may employ a linear combination to generate the updated KB:
\begin{equation}
\label{eq:p3}
  P_{k+1}(c_i)
	= \alpha_k  P_{k}(c_i) + \beta_k(A) P_k(c_i\mid A),\; i=1,\ldots,5,
\end{equation} 
where $\alpha_k+\beta_k(A)=1$. Here, one may utilize $\beta_k(A)$, perhaps as $\beta_k(A)=0.75$ to quantify the confidence on new evidence assigned by the agent $\mc{A}^*$. One may interpret accordingly and use the parameter $\alpha$ to account for the integrity of existing KB.

\paragraph{Belief theoretic update}
Similar to the probabilistic case, When a conditioning event, such as $A=\Theta\setminus\{c_1,c_2\}=\{c_3,c_4,c_5\}$ (i.e., $c_1$ and $c_2$ are no longer in the running) is specified with 100\% certainty, one can interpret this as the original FoD being deflated (i.e., the \# of available candidates are being reduced) to $(c_3,c_4,c_5)$. Therefore, we can generate updated KB via FH Conditional as
\begin{equation}
\label{eq:b1}
  \Bl_{k+1}(B)
	= \Bl_k(B\mid A),
\end{equation}
where $B\subseteq\Theta$. Similar to the probabilistic case, the conditioning operation, redistributes originally cast voter preferences to ONLY propositions that are subsets of $A$. Now, when $A$ is specified with less than 100\% confidence as in Example~\ref{ex:e_poll2}, one may utilize updating strategy similar to~\cite{Kul04} to obtain,
\begin{equation}
\label{eq:b2}
  \Bl_{k+1}(B)
	= \alpha_k \Bl_{k}(B) + \beta_k(A) \Bl_k(B\mid A),
\end{equation}
where the parameters $\alpha_k$ and $\beta_k(A)$ are chosen to reflect the confidence levels. In particular, for Example~\ref{ex:e_poll2}, one may choose $\beta_k(A)=0.75$ and $\alpha_k=1-\beta_k(A)=0.25$. An interpretation of $\alpha_k$ is that it represents the integrity of current KB (i.e., $\Bl_{k}(\cd)$) in front of not 100\% evidence. In fact, depending on the maturity of the KB, one may choose an $\alpha_k$ not fully committing to incoming evidence (see~\cite{Kul04} for a detailed discussion). For instance, even if $\mc{A}^*$ is 95\% confident in his assessment on $c_1$ and $c_2$, if the knowledge in existing KB has much higher integrity, one may choose a higher $\alpha_k$, say $0.80$, thus only allowing small changes (20\% in a very loose sense) to the existing KB. 

\subsection{The Case of Two Disjoint Conditioning Events}
In complex fusion environments, especially when evidence is pooled from open sources, such as in crowd-sensing applications, it is highly unlikely that incoming evidence constitutes of a single event along with a confidence value. Let us look at a simplified scenario, where the incoming evidence constitutes ONLY of two disjoint events.

\begin{example}[MVP Poll v.3]
\label{ex:e_poll3}
Suppose the sports analyst $\mc{A}^*$ in Example~\ref{ex:e_poll2} has now gathered more evidence regarding the condition of  $c_1$ and $c_2$. Now, he's 100\% certain that either event $A_1$:= $c_1$ and $c_2$ will not return with 90\% chance, or otherwise event $A_2$:= if they return, due to their utmost dedication, only $c_1$ and $c_2$ will have a chance at the MVP. \bi{Now, what is the best strategy to update $\mc{E}_k$?}
\end{example}

\paragraph{Probabilistic Case}
Let us assume that current voter preferences are stored in KB as $P_k(\cd)$, where $k$ is the discrete time index for current time $t_k$. Now, the evidence is provided via two disjoint events $A_1=(c_3,c_4,c_5)$ and $A_2=(c_1,c_2)$. For a proposition $B\subseteq\Theta$, one may compute 
\begin{equation}
\label{eq:p4}
P_{k}(B) = P_{k}(B|A_1)P_k(A_1) + P_{k}(B|A_2)P_k(A_2)
\end{equation}
Perhaps, one direct way to extend this notion of total probability to derive up update equation similar to equation~\eqref{eq:p4}, while taking into account the less than perfect confidence of incoming evidence and it's impact on the integrity of the KB (i.e., $P_{k}(B)$). Therefore, one may derive the intuitive extension, similar to Eq.~\eqref{eq:b2}, as
\begin{align}\notag
\label{eq:p5}
P_{k+1}(B) 	& = \alpha_k P_{k}(B)+ (1-\alpha)[P_{k}(B|A_1)P^*_k(A_1)\\\notag
			& \qquad\qquad\qquad\qquad\qquad+ P_{k}(B|A_2)P^*_k(A_2)\\
			&=  \alpha_k P_{k}(B) + \sum_{i=1}^2\beta_k(A_i) P_{k}(B|A_i),
\end{align}
where $\beta_k(A_i)=(1-\alpha_k)P^*_k(A_i)$ with $\alpha_k+\sum_i \beta_k(A_i)=1$. Here, note that the support for each conditioning event $A_i$ as provided by $\mc{E}^*_k$ is used. Perhaps, one may interpret this as a weighted linear combination of conditioned evidence, where the parameters $\beta_k(A_i)$ are directly proportional to the support provided by incoming evidence to the conditioning events $A_i,\,i=1,2$ as given by $\beta_k(A_i)=(1-\alpha_k)P^*_k(A_i)$.

\paragraph{Belief theoretic update}
Now, one may directly extend the same thought process in Eq.~\ref{eq:p5} as 
\begin{align}
\label{eq:b4}
  \Bl_{k+1}(B)  = \alpha_k\Bl_{k}(B) + \sum_{i=1}^2\beta_k(A_i)\Bl_k(B\mid A_i),
\end{align}
where $\alpha_k+\sum_i \beta_k(A_i)=1$. Here, the parameters $\beta_k(A_i)$ to be chosen s.t. they are directly proportional to the support for event $A_i$ as provided by $\mc{E}^*_k$.

\subsection{The Case of Arbitrary (Multiple) Conditioning Events}
The most general case can be analyzed when the incoming evidence $\mc{E}^*_k$ is expressed as a BoE $\mc{E}^*_k\equiv\{\Theta,\mf{F}^*_k,m^*_k(\cd)\}$, where the focal set $\mf{F}^*_k$ and basic probability assignment $m^*_k(\cd)$ containing the set of conditioning events and their support, respectively.

Following the notion of generating updated belief as a weighted linear combination of conditioned evidence, where the parameters $\beta_k(A_i)$ are taken to be directly proportional to the support provided by incoming evidence, one can extend Eq.~\ref{eq:b4} for updating with arbitrary conditioning events.

\begin{definition}[Generalized Conditional Update (GCU)]
\label{def:GCU}
The GCU that updates $\mc{E}_k\equiv\{\Theta,\mf{F}_k,m_k(\cd)\}$ with the evidence in $\mc{E}_k^*\equiv\{\Theta,\mf{F}_k^*,m_k^*(\cd)\}$ is given by
\begin{align}
\label{eq:b4}
  \Bl_{k+1}(B)  = \alpha_k\Bl_{k}(B) + \sum_{A\in\mf{F}^*_k}\beta_k(A)\Bl_k(B\mid A),
\end{align}
where the $\alpha_k, \beta_k(\cd)\in\Re^+$ satisfy $\alpha_k + \sum\limits_{A\in\mf{F}^*_k}\beta_k(A)=1$.
\end{definition}

\noindent
\bi{Remarks:}
The updating equations given by GCU proposed in this paper and CUE in~\cite{Pre09_ICIF} have similar functional form. However, the updating schemes are fundamentally different. In particular, 

	--- GCU \ul{conditions existing evidence} $\mc{E}_k$ w.r.t. conditioning events $A\in\mf{F}^*_k$ provided by incoming evidence in $\mc{E}_k^*$, whereas 
	
	--- CUE \ul{conditions incoming evidence} $\mc{E}_k^*$ within itself w.r.t. $A\in\mf{F}^*_k$.

This fundamental difference generates interesting differences for evidence updating and provides different and distinct features that maybe relevant in certain application contexts. In fact, we show that GCU boils down to Bayes conditional under the limiting conditions (i.e., $\alpha_k=0,\forall k$ and $m_{k=0}(\cd)$ is Bayesian), irrespective of the structure of incoming evidence $\mc{E}_k$. These features are directly influenced by the properties of FH conditional operation.

\section{Behavior of GCU}

Understanding how the conditioning affects the focal elements in the current BoE is crucial to a proper understanding of any updating process.

\subsection{Focal Elements Generated via Conditioning}
 A theorem that explains the focal elements generated by conditioning referred to as \emph{Conditional Core Theorem}~\cite{Wic10_ICIF,Wic13ToSMCB} can be utilized for this task.

\begin{theorem}[Conditional Core Theorem (CCT)]
\label{thm:cond_focs}
\cite{Wic10_ICIF}
Given $A\in\hat{\mf{F}}$ in the BoE $\mc{E}=\{\Theta, \mf{F}, m\}$, $m(B|A)>0$ iff $B$ can be expressed as $B=X\cup Y$, for some $X\in\tx{in}(A)$ and $Y\in\tx{OUT}(A)\cup\{\emptyset\}$. Here, $\tx{in}(A)=\{B\subseteq A\mid B\in\mf{F}\}$, $\tx{OUT}(A)=\{B\subseteq A\mid B=\bigcup_{i\subseteq\mc{I}} C_i,\,C_i\in\tx{out}(A)\}$, where $\tx{out}(A)=\{B\subseteq A\mid B\cup C\in\mf{F},\,\emptyset\neq B,\,\emptyset\neq C\subseteq\ol{A}\}$.
\endproof
\end{theorem}

The CCT implies that conditional focal elements can only be generated from the disjunction of focal elements completely contained in the conditioning proposition $A$ and focal elements that intersect but not included in $A$. For a comprehensive discussion on CCT, we refer the interested reader to \cite{Wic13ToSMCB}. The following example~\cite{Wic10_ICIF} illustrates the application of the CCT.

\begin{example} 
\label{ex:cct}
\cite{Wic10_ICIF}
Consider the BoE, $\mc{E}_k\equiv\{\Theta,\mf{F}_k,m_k(\cd)\}$ with $\Theta=\{a, b, c, d, e, f, g, h, i\},\,m_k=\{a, b, h, df, beg, \Theta\}$ and $m_k(B)=\{0.1, 0.1, 0.1, 0.2, 0.2, 0.3\}$, for $B\in \mf{F}_k$ (in the same order given in $\mf{F}_k$). Then, for $A=(abcde)$, 
\begin{alignat*}{4}
  &\tx{in}(A)
    &
      &=\{a, b\};\;
        &
          &\tx{out}(A)
            &
              &=\{d, be, abcde\}; \\
  &\tx{IN}(A)
    &
      &=\{a, b, ab\};\;
        &
          &\tx{OUT}(A)
            &
              &=\{d, be, bde, abcde\}.
\end{alignat*}
Note that, $\mc{B}=\{ad, bd, be, abe, bde, abde, abcde\}$, are the only propositions that can be expressed as $B=X\cup Y$, for some $X\in\tx{in}(A)$ and $Y\in\tx{OUT}(A)$. So, according to the CCT, the nine elements of $\mc{B}$ and $\tx{in}(A)$ are the only propositions that will belong to the conditional core (w.r.t. $A=(abcde)$). 
\endproof
\end{example}

\begin{figure}[htb!]
\centering
\includegraphics[width=3.5in]{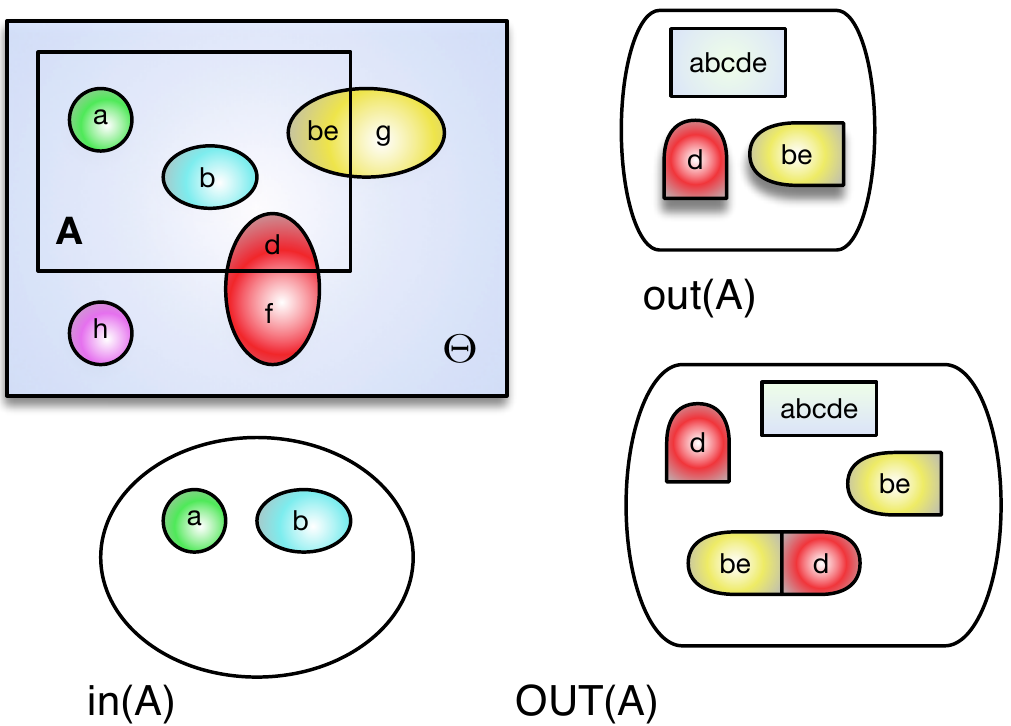}
\vspace*{-0.1in}
\caption{Illustration of the application of CCT for conditioning $\mc{E}_k=\{\Theta,\mf{F}_k,m_k(\cd)\}$ with $\Theta=\{a, b, c, d, e, f, g, h, i\},\,m=\{a, b, h, df, beg, \Theta\}$ and $m(B)=\{0.1, 0.1, 0.1, 0.2, 0.2, 0.3\}$ w.r.t. the conditioning event $A=(abcde)$.}
\label{fig:ex:cct}
\vspace*{-0.1in}
\end{figure}

\subsection{Impact on the Updating Process}
The conditioning operation has a direct impact on the updating process. For pedagogical ease, let us consider the update with $\alpha_k=0$, i.e., $\mc{E}_{k+1}$ will only contain focal elements that are generated via conditioning. 

  (a)~If proposition $B$ was not contained in at least one the conditioning events $A\in\mf{F}^*_k$ (i.e., $B\not\subseteq A,\,\forall A\in\mf{F}^*_k$), then $B$ cannot belong to $\mf{F}_{k+1}$.
  
  (b)~For a proposition $B$ that was contained in at least one conditioning event $A\in\mf{F}^*_k$,
  
  \hspace*{0.1in}(b.1)~if $B\in\mf{F}_k$ (i.e., it is a focal element in $\mc{E}_k$), then $B$ belongs to $\mf{F}_{k+1}$.
  
  \hspace*{0.1in}(b.2)~if $B\not\in\mf{F}_k$, then $B$ belongs to $\mf{F}_{k+1}$ \bi{iff} it can be expressed as the union of \\
	\hspace*{0.3in} (i) a focal element $\hat{B}\in\mf{F}_k$ that is contained in a conditioning event $A\in\mf{F}^*_k$; and \\
	\hspace*{0.3in} (ii) the intersection of $A$ with some arbitrary set of focal elements $\tilde{B}\in\mf{F}_k$ each of which straddles $A$ and its complement $\ol{A}$ (see Fig.~\ref{fig:ex:cct}).

These properties clearly identifies how new propositions will be generated or existing propositions will be removed via conditioning operations. Furthermore, 

  (d)~for each conditioning operation, if there are no focal elements that straddle any of the conditioning events $A\in\mf{F}^*_k$ and its complement, then $\mf{F}_{k+1}=\mf{F}_k$;
  
  (e)~propositions with zero belief does not belong to $\mf{F}_{k+1}$.

Therefore, with the exception of newly created focal elements (from straddling propositions as explained about), when $\alpha=0$ (i.e., completely update existing BoE with incoming evidence as in Bayes conditioning), the FH conditioning in GCU eliminates all propositions that were in complete disagreement with ALL conditioning events (as in the Bayes Conditioning). 

Now, when $\alpha_k>0$, the core $\mf{F}_{k+1}$ of updated BoE $\mc{E}_{k+1}$ also retains all the focal elements that were contained in $\mf{F}_{k}$. Since $\mf{F}_{k+1}$ retains all focal elements that were contained in at least one of the conditioning events $A\in\mf{F}_k^*$ (as is the case in Bayes Conditioning), it is interesting to look at the asymptotic behavior of propositions that are not supported by incoming evidence. Furthermore, how one initiates the updating process, or in other words, the initial mass assignments or the priors, will also have a clear impact on the final updated BoEs. In particular, it is easy to see that the vacuous initial assignment (i.e., $m_{k=0}(\Theta)=1.0$) as a representation of complete ambiguity would not generate any refined BoEs irrespective of the incoming evidence. 

\subsection{Vacuous Updating}
Let $\mc{E}_{\Theta}$ denote the vacuous BoE, i.e., the complete ambiguity often represented by $\mc{E}_{\Theta}=\{\Theta, \Theta, m(\Theta)=1.0\}$.

(i) Case 1: $\mc{E}_{k+1}:=\mc{E}_{\Theta}\lhd\mc{E}^*_k$, i.e., updating a vacuous BoE with an arbitrary BoE:
	GCU generates $\mc{E}_{k+1}=\mc{E}_{\Theta}$, irrespective of parameter selection. In particular, if one initiates an updating process with $\mc{E}_{k=0}=\mc{E}_{\Theta}$, perhaps due to lack of prior information, $\mc{E}_{k+1}:= \mc{E}_{k}\lhd\mc{E}^*_k=\mc{E}_{\Theta}$, for all $k$. In other words, the KB continues to remain vacuous irrespective of the parameter selection or the incoming evidence. Obviously, conditioning only generates a refinement of originally cast evidence, and therefore proper selection of initial basic probability assignment (or priors) is crucial with the use of GCU. 

(ii) Case 2: $\mc{E}_{k+1}:=\mc{E}_k\lhd\mc{E}_{\Theta}$, i.e., updating an arbitrary BoE with a vacuous BoE:
GCU generates $\mc{E}_{k+1}=\mc{E}_{k}$, again, irrespective of parameter selection. In particular, if one carries out $n$ updates as $\mc{E}_{i+1}:=\mc{E}_i\lhd\mc{E}_{\Theta}$, for $i=k,\ldots,k+n-1$, then, irrespective of GCU parameters, $\mc{E}_{k+n}:=\mc{E}_k$. Therefore, GCU updates are tolerant against complete sensor failures in the sense that it will not complete erode an existing KB with vacuous incoming evidence.

\subsection{Selection of GCU Parameters}
GCU provides flexibility in parameter selection in order to accommodate the integrity of the existing knowledge-base, reliability of sensors and other application specific requirements. How one goes about selecting the appropriate parameter configuration however is is highly dependent on the application and domain. The work in \cite{Kul04, Pre09_ICIF, Wic10_WTBF} details several parameter selection strategies for CUE-based evidence updating. These strategies remain applicable for GCU. We do not intend to repeat a detailed description of these strategies here; the interested reader may refer to \cite{Kul04, Pre07, Wic10_WTBF}. 

\paragraph{Selection of $\alpha_k$}
The work in~\cite{Kul04} provides several strategies for selection of $\alpha_k$ w.r.t. the ``inertia of existing body of evidence.'' In particular, 
(i)~\emph{infinite inertia}-based selection: $\alpha_k=1$;
(ii)~\emph{zero inertia}-based selection: $\alpha_k=0$; and, 
(iii)~\emph{proportional inertia}-based selection: k/(k+1), where $k$ current discrete time index.

\paragraph{Selection of $\beta_k(\cd)$} 
These weights allow one to emphasize/de-emphasize the propositions within each conditioning set $A$. Two very interesting choices that are inspired by the work in \cite{Kul04, Pre07, Wic10_WTBF} are the following:

(i) The receptive strategy:
$\beta_k(A)=K_km_k^*(A),\,\forall A\in\mf{F}_k^*$, where $K_k\neq 0$ is a constant. This receptive strategy `weighs' the incoming evidence from $\mc{E}_k^*$ according to the support $\mc{E}_k^*$ 
itself has for it. In other words, the BoE that is being updated is `receptive' to the support (assigned to the conditioning events) from  incoming evidence $\mc{E}_k^*$.
  
(ii) The cautious strategy:
$\beta_k(A)=K_km_k(A),\,\forall A\in\mf{F}_k$, where $K_k\neq 0$ is a constant. This strategy `weights' the incoming evidence from $\mc{E}_k^*$ according to the support $\mc{E}_k$ itself has for it. In other words, being KB is being `cautious' as to what events are in fact used for conditioning (or refining).

\subsection{Selection of Priors or Initial Mass Assignment}
DS theoretic sensor fusion methods are perhaps popular for its ability to conveniently ambiguous data, especially ignorance. For instance, with lack of information, one may start a fusion process by initializing $\mc{E}_0$ with $m_0(\Theta)=1.0$, whereas in Bayesian probability, one may choose an approach, such as uniform priors. Unlike combination operators, since conditioning generates a refinement by redistributing initial mass assignments within the conditioning event(s), proper selection of initial mass assignment is crucial with GCU. 

\paragraph{Selecting a vacuous BoE for $\mc{E}_0$}
As shown, the uddate $\mc{E}_{k+1}:=\mc{E}_{\Theta}\lhd\mc{E}^*_k$ generates $\mc{E}_{k+1}=\mc{E}_{\Theta}$. Therefore, irrespective of $\mc{E}^*_k$, vacuous initial mass assignment will result in $\mc{E}_{k\to\infty}=\mc{E}_{\Theta}$.

\paragraph{Selecting a uniform mass assignment for $\mc{E}_0$}
As in probability theory, one may utilize a uniform assignment, as $m_0(\theta_i)=1/|\Theta|,\,i=1,\ldots,|\Theta|$. 

\paragraph{Selecting a Dirichlet BoE for $\mc{E}_0$}
Another, perhaps more `DS like' assignment strategy is to utilize, a Dirichlet BoE. With an application/context related parameter $0<\gamma<1$, one may initialize the updating process as, $m_0(\theta_i)=(1-\gamma)/|\Theta|,\,i=1,\ldots,|\Theta|$ and $m_0(\Theta)=\gamma$. Perhaps, the gamma maybe chosen to represent the `ignorance' associated with initial assignment.

\section{Concluding Remarks}
\label{sec:conc}

Robust belief revision methods are crucial in streaming data situations for updating existing knowledge or beliefs with new incoming evidence. By viewing the evidence updating process as a thought experiment, a novel evidence updating strategy referred to as the GCU (i.e., generalized conditional update) is derived, especially targeting efficient belief revision and uncertainty handling in big-data stream-processing applications. The GCU generalizes the belief theoretic notion of Fagin-Halpern conditional, thus allowing one to account for various data/source imperfections that are characteristic to complex big-data fusion environments. The presented extension differs fundamentally from the previously published work on the conditional approach referred to as CUE (i.e., Conditional Update Equation) as well as authors own extensions of it. The GCU, the proposed evidence updating strategy based on a belief theoretic conditional approach, possesses several intuitively appealing features which seem to indicate its suitability for scenarios where large amounts of data/source uncertainties are present. In particular, the initial basic belief assignment (or the priors) is fundamentally different from CUE and it's counterparts. As shown in the derivations, a vacuous assignment as prior or initial basic probability assignment (as often done in Dempster-Shafer settings) is shown to generate inaccurate results. Among issues that warrant further investigation, of particular importance is the computational complexity that hampers the use of DS theoretic methods when working with a high number of sources and/or source FoDs having high cardinality.

\section*{Acknowledgment}
The author would like to thank Prof. Kamal Premaratne (University of Miami, Coral Gables) for enlightening discussions on numerous related topics and reviewers for their invaluable feedback.

\bibliographystyle{../../../bib/IEEEtran}
\bibliography{%
../../../bib/IEEEabrv,%
../../../bib/articles_journals,%
../../../bib/articles_conferences,%
../../../bib/articles_books,%
../../../bib/articles_urls,%
../../../bib/articles_other}

\begin{thebibliography}{10}
\providecommand{\url}[1]{#1}
\csname url@rmstyle\endcsname
\providecommand{\newblock}{\relax}
\providecommand{\bibinfo}[2]{#2}
\providecommand\BIBentrySTDinterwordspacing{\spaceskip=0pt\relax}
\providecommand\BIBentryALTinterwordstretchfactor{4}
\providecommand\BIBentryALTinterwordspacing{\spaceskip=\fontdimen2\font plus
\BIBentryALTinterwordstretchfactor\fontdimen3\font minus
  \fontdimen4\font\relax}
\providecommand\BIBforeignlanguage[2]{{%
\expandafter\ifx\csname l@#1\endcsname\relax
\typeout{** WARNING: IEEEtran.bst: No hyphenation pattern has been}%
\typeout{** loaded for the language `#1'. Using the pattern for}%
\typeout{** the default language instead.}%
\else
\language=\csname l@#1\endcsname
\fi
#2}}

\bibitem{Val09}
E.~D. Valle, S.~Ceri, F.~v.~Harmelen, and D.~Fensel, ``It's a streaming world!
  reasoning upon rapidly changing information,'' \emph{{IEEE} {I}ntelligent
  {S}ystems}, vol.~24, no.~6, pp. 83--89, Nov 2009.

\bibitem{Esp15}
C.~Esposito, M.~Ficco, F.~Palmieri, and A.~Castiglione, ``A knowledge-based
  platform for big data analytics based on publish/subscribe services and
  stream processing,'' \emph{Knowledge-Based Systems}, vol.~79, pp. 3 -- 17,
  2015.

\bibitem{jams16}
\BIBentryALTinterwordspacing
P.~Jamshidi and G.~Casale, ``{An Uncertainty-Aware Approach to Optimal
  Configuration of Stream Processing Systems},'' June 2016. [Online].
  Available: \url{https://doi.org/10.5281/zenodo.56238}
\BIBentrySTDinterwordspacing

\bibitem{Li_ToBD16}
T.~Li, J.~Tang, and J.~Xu, ``Performance modeling and predictive scheduling for
  distributed stream data processing,'' \emph{IEEE Transactions on Big Data},
  vol.~2, no.~4, pp. 353--364, Dec 2016.

\bibitem{Wang_PRL17}
Y.~Wang, H.~Gao, and G.~Chen, ``Predictive complex event processing based on
  evolving bayesian networks,'' \emph{Pattern Recognition Letters}, 2017.

\bibitem{Kir03}
T.~Kirubarajan and Y.~Bar-Shalom, ``{Kalman} filter versus {IMM} estimator:
  {W}hen do we need the latter?'' \emph{{IEEE} Transactions on Aerospace and
  Electronic Systems}, vol.~39, no.~4, pp. 1452--1457, Oct. 2003.

\bibitem{Hal92_book}
D.~L. Hall, \emph{Mathematical Techniques in Multisensor Data Fusion}.\hskip
  1em plus 0.5em minus 0.4em\relax Norwood, MA: Artech House, 1992.

\bibitem{Wic12_ICBF}
T.~L. Wickramarathne, K.~Premaratne, and M.~N. Murthi, ``Consensus-based
  credibility estimation of soft evidence for robust data fusion,'' in
  \emph{Belief Functions}, ser. Advances in Intelligent and Soft Computing,
  T.~Denoeux and M.-H. Masson, Eds.\hskip 1em plus 0.5em minus 0.4em\relax
  Compi\'egne, France: Springer Berlin / Heidelberg, May 2012, vol. 164, pp.
  301--309.

\bibitem{Lam94}
W.~Lam and F.~Bacchus, ``Learning {B}ayesian belief networks: {A}n approach
  based on the {MDL} principle,'' \emph{Computational Intelligence}, vol.~10,
  pp. 269--293, July 1994.

\bibitem{Hec95}
D.~Heckerman, D.~Geiger, and D.~Chickering, ``Learning {B}ayesian networks:
  {T}he combination of knowledge and statistical data,'' \emph{Machine
  Learning}, vol.~20, no.~3, pp. 197--243, 1994.

\bibitem{Fri97b}
N.~Friedman, D.~Geiger, and M.~Goldszmidt, ``{Bayes}ian network classifiers,''
  \emph{Machine Learning}, vol.~29, no. 2/3, pp. 131--163, 1997.

\bibitem{Jen01_book}
F.~V. Jensen, \emph{Bayesian Networks and Decision Graphs}, 1st~ed.\hskip 1em
  plus 0.5em minus 0.4em\relax New York, NY: Springer-Verlag, 2001.

\bibitem{Hua02}
H.-J. Huang and C.-N. Hsu, ``{Bayes}ian classification for data from the same
  unknown class,'' \emph{{IEEE} Transactions on Systems, Man and Cybernetics,
  Part {B}: Cybernetics}, vol.~32, no.~2, pp. 137--145, Apr. 2002.

\bibitem{Man05}
S.~Mani, M.~Valtorta, and S.~McDermott, ``Building {B}ayesian network models in
  medicine: The {MENTOR} experience,'' \emph{Applied Intelligence}, vol.~22,
  no.~2, pp. 93--108, 2005.

\bibitem{Fag91_UAI}
R.~Fagin and J.~Y. Halpern, ``A new approach to updating beliefs,'' in
  \emph{Proc. Conference on Uncertainty in Artificial Intelligence ({UAI})},
  P.~P. Bonissone, M.~Henrion, L.~N. Kanal, and J.~F. Lemmer, Eds.\hskip 1em
  plus 0.5em minus 0.4em\relax New York, NY: Elsevier Science, 1991, pp.
  347--374.

\bibitem{Ken92}
R.~Kennes, ``Computational aspects of the {M\"{o}}bius transformation of
  graphs,'' \emph{{IEEE} Transactions on Systems, Man and Cybernetics},
  vol.~22, no.~2, pp. 201--223, Mar. 1992.

\bibitem{Chr95}
L.~Chrisman, ``Incremental conditioning of lower and upper probabilities,''
  \emph{International Journal of Approximate Reasoning}, vol.~13, no.~1, pp.
  1--25, July 1995.

\bibitem{Abi92_book}
M.~A. Abidi and R.~C. Gonzalez, \emph{Data Fusion in Robotics and Machine
  Intelligence}.\hskip 1em plus 0.5em minus 0.4em\relax San Diego, CA: Academic
  Press, 1992.

\bibitem{Mitchell12_book}
H.~B. Mitchell, \emph{Data Fusion: Concepts and Ideas}, 2nd~ed.\hskip 1em plus
  0.5em minus 0.4em\relax {Springer}, 2012.

\bibitem{Bog87}
P.~L. Bogler, ``Shafer-{D}emspter reasoning with applications to multisensor
  target identification systems,'' \emph{{IEEE} Transactions on Systems, Man
  and Cybernetics}, vol.~17, no.~6, pp. 968--977, 1987.

\bibitem{Pea88_book}
J.~Pearl, \emph{Probabilistic Reasoning in Intelligent Systems: Networks of
  Plausible Inference}.\hskip 1em plus 0.5em minus 0.4em\relax San Francisco,
  CA: Morgan Kaufmann, 1988.

\bibitem{Elo01}
Z.~Elouedi, K.~Mellouli, and P.~Smets, ``Belief decision trees: theoretical
  foundations,'' \emph{International Journal of Approximate Reasoning},
  vol.~28, no. 2/3, pp. 91--124, Nov. 2001.

\bibitem{Del04}
F.~Delmotte and P.~Smets, ``Target identification based on the transferable
  belief model interpretation of {D}empster-{S}hafer model,'' \emph{{IEEE}
  Transactions on Systems, Man and Cybernetics, Part {A}: Systems and Humans},
  vol.~34, no.~4, pp. 457--471, July 2004.

\bibitem{Che05}
T.~M. Chen and V.~Venkataramanan, ``{D}emspter-{S}hafer theory for intrusion
  detection in ad hoc networks,'' \emph{{IEEE} Internet Computing}, vol.~9,
  no.~6, pp. 35--41, Nov./Dec. 2005.

\bibitem{Ris05}
B.~Ristic and P.~Smets, ``Target identification using belief functions and
  implication rules,'' \emph{{IEEE} Transactions on Aerospace and Electronic
  Systems}, vol.~41, no.~3, pp. 1097--1103, July 2005.

\bibitem{Den06_SMC}
T.~Denoeux and P.~Smets, ``Classification using belief functions: the
  relationship between the case-based and model-based approaches,''
  \emph{{IEEE} Transaction on Systems, Man and Cybernetics, Part {B}:
  Cybernetics}, vol.~36, no.~6, pp. 1395--1406, 2006.

\bibitem{Mar06}
A.~Martin and C.~Osswald, ``Human experts fusion for image classification,''
  \emph{Information and Security: An International Journal, Special Issue on
  Fusing Uncertain, Imprecise and Conflicting Information}, vol.~20, pp.
  122--141, May 2006.

\bibitem{Mas11}
M.~H. Masson and T.~Denoeux, ``Ensemble clustering in the belief functions
  framework,'' \emph{International Journal of Approximate Reasoning}, vol.~52,
  no.~1, pp. 92--109, Jan. 2011.

\bibitem{Bur06}
J.~Burke, D.~Estrin, M.~Hansen, A.~Parker, N.~Ramanathan, S.~Reddy, and M.~B.
  Srivastava, ``Participatory sensing,'' in \emph{Proc. Workshop on
  World-Sensor-Web ({WSW}): Mobile Device Centric Sensor Networks and
  Applications}, 2006, pp. 117--134.

\bibitem{Hall12_book}
D.~Hall, M.~L. II, C.-Y. Chong, and J.~Linas, \emph{Distributed Data Fusion for
  Network-Centric Operations}.\hskip 1em plus 0.5em minus 0.4em\relax Boca
  Raton, FL: {CRC} Press, 2012.

\bibitem{Sha76_book}
G.~Shafer, \emph{A Mathematical Theory of Evidence}.\hskip 1em plus 0.5em minus
  0.4em\relax Princeton, NJ: Princeton University Press, 1976.

\bibitem{Pre07}
K.~Premaratne, D.~A. Dewasurendra, and P.~H. Bauer, ``Evidence combination in
  an environment with heterogeneous sources,'' \emph{{IEEE} Transactions on
  Systems, Man and Cybernetics, Part {A}: Systems and Humans}, vol.~37, no.~3,
  pp. 298--309, 2007.

\bibitem{Pre09_ICIF}
K.~Premaratne, M.~N. Murthi, J.~Zhang, M.~Scheutz, and P.~H. Bauer, ``A
  {D}empster-{S}hafer theoretic conditional approach to evidence updating for
  fusion of hard and soft data,'' in \emph{Proc. International Conference on
  Information Fusion ({FUSION})}, Seattle, {WA}, July 2009, pp. 2122--2129.

\bibitem{Wic10_WTBF}
T.~L. Wickramarathne, K.~Premaratne, M.~N. Murthi, and M.~Scheutz, ``A
  {D}empster-{S}hafer theoretic evidence updating strategy for non-identical
  frames of discernment,'' in \emph{Proc. Workshop on the Theory of Belief
  Functions ({BELIEF})}, Brest, France, Apr. 2010.

\bibitem{Wic13_ICASSP}
T.~L. Wickramarathne, K.~Premaratne, and M.~N. Murthi, ``Convergence analysis
  of consensus belief functions within asynchronous ad-hoc fusion networks,''
  in \emph{Proc. International Conference on Statistical Signal Processing
  ({ICASSP})}, Vancouver, {C}anada, May 2013.

\bibitem{Kul04}
E.~C. Kulasekere, K.~Premaratne, D.~A. Dewasurendra, M.-L. Shyu, and P.~H.
  Bauer, ``Conditioning and updating evidence,'' \emph{International Journal of
  Approximate Reasoning}, vol.~36, no.~1, pp. 75--108, Apr. 2004.

\bibitem{Sme99_UAI}
P.~Smets, ``Practical uses of belief functions,'' in \emph{Proc. Conference on
  Uncertainty in Artificial Intelligence ({UAI})}, K.~B. Laskey and H.~Prade,
  Eds.\hskip 1em plus 0.5em minus 0.4em\relax San Francisco, CA: Morgan
  Kaufmann, 1999, pp. 612--621.

\bibitem{Wic10_ICIF}
T.~L. Wickramarathne, K.~Premaratne, and M.~N. Murthi, ``Focal elements
  generated by the {D}empster-{S}hafer theoretic conditionals: A complete
  characterization,'' in \emph{Proc. International Conference on Information
  Fusion ({FUSION})}, Scotland, {UK}, July 2010.

\bibitem{Wic13ToSMCB}
------, ``Toward efficient computation of the dempster-shafer belief theoretic
  conditionals,'' \emph{{IEEE} Transactions on Cybernetics}, vol.~43, no.~2,
  pp. 712--724, Apr. 2012.

\end{thebibliography}

\end{document}